\documentclass[letterpaper, 10 pt, conference]{ieeeconf}  

\usepackage{amsmath,amsfonts}
\usepackage{algorithmic}
\usepackage{algorithm}
\usepackage{array}
\usepackage{subcaption}

\usepackage{graphicx}
\usepackage{tabularx}
\usepackage[dvipsnames]{xcolor}
\usepackage{url}

\usepackage{booktabs}       

\let\labelindent\relax
\usepackage{enumitem}

\usepackage{arydshln}
\makeatletter
\def\adl@drawiv#1#2#3{%
        \hskip.5\tabcolsep
        \xleaders#3{#2.5\@tempdimb #1{1}#2.5\@tempdimb}%
                #2\z@ plus1fil minus1fil\relax
        \hskip.5\tabcolsep}
\newcommand{\cdashlinelr}[1]{%
  \noalign{\vskip\aboverulesep
           \global\let\@dashdrawstore\adl@draw
           \global\let\adl@draw\adl@drawiv}
  \cdashline{#1}
  \noalign{\global\let\adl@draw\@dashdrawstore
           \vskip\belowrulesep}}
\makeatother

\IEEEoverridecommandlockouts                              

\def\papertitle{Pick Planning Strategies for Large-Scale Package Manipulation}

\def\publishedurl{\centering \textbf{{\color{red}{Learning Meets Model-based Methods for Manipulation and Grasping Workshop}}\\
{\color{red}{2023 IEEE/RSJ International Conference on Intelligent Robots and Systems (IROS), October 1 - 5, 2023}}\\
\color{blue}{\url{https://sites.google.com/view/learning-meets-models-iros2023}}}}

\title{\LARGE \bf
\papertitle
}

\author{Shuai Li$^{1}$, Azarakhsh Keipour$^{1}$, Kevin Jamieson$^{1,2}$, Nicolas Hudson$^{1}$,\\Sicong Zhao$^{1}$, Charles Swan$^{1}$ and Kostas Bekris$^{1,3}$
\thanks{$^{1}$Amazon Robotics, United States
        {\tt\small \{amzshua, keipourv, jamikevi, hudnco, sczhao, cswan, bekris\}@amazon.com}}%
\thanks{$^{2}$Allen School of Computer Science \& Engineering, University of Washington, Seattle, Washington 98105, USA}%
\thanks{$^{3}$School of Arts and Sciences, Rutgers University, Piscataway, New Jersey 08854, USA}%
}

\begin{document}

\maketitle
\thispagestyle{empty}
\pagestyle{empty}

\begin{abstract}
Automating warehouse operations can reduce logistics overhead costs, ultimately driving down the final price for consumers, increasing the speed of delivery, and enhancing the resiliency to market fluctuations. 

This extended abstract showcases a large-scale package manipulation from unstructured piles in Amazon Robotics' Robot Induction (Robin) fleet, which is used for picking and singulating up to 6~million packages per day and so far has manipulated over 2~billion packages. It describes the various heuristic methods developed over time and their successor, which utilizes a pick success predictor trained on real production data.

To the best of the authors' knowledge, this work is the first large-scale deployment of learned pick quality estimation methods in a real production system.

\end{abstract}


\section{Introduction} \label{sec:intro}

Automation in the industrial, manufacturing, and warehouse sectors holds immense potential for reducing overhead expenses related to the production, handling, and sorting goods. By enabling higher speed and precision in product handling, it has the capacity to lower customer costs and enhance product quality. Additionally, automation helps mitigate risks to human workers involved in manual operations while bolstering resilience against economic fluctuations.

Tasks - such as picking objects from unstructured, cluttered piles - have only recently become robust enough for large-scale deployment with minimal human intervention. This extended abstract intends to showcase the efforts in deploying a large-scale package manipulation from unstructured piles by Amazon Robotics' Robot Induction (Robin) fleet~\cite{robin}. The progression of the methods developed over time varies from simple heuristic algorithms to utilizing a pick success predictor trained on real production data~\cite{Li:2023:rss:pickranking}. The system is currently being used for singulating up to 6~million packages per day and so far has manipulated over 2~billion packages.

\begin{figure}[t]
    \centering
    \includegraphics[width=0.497\linewidth, height=5cm]{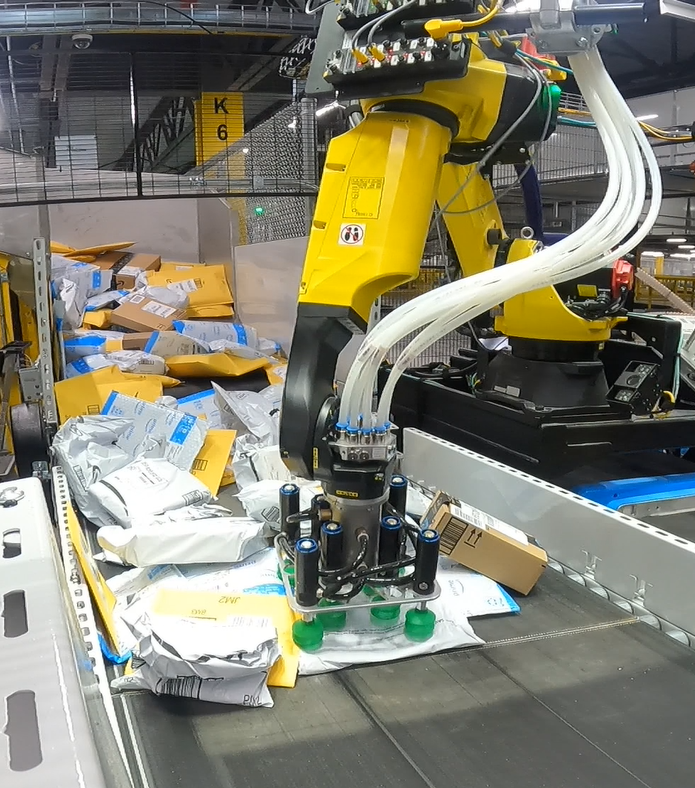}%
    \hfill%
    \includegraphics[width=0.497\linewidth, height=5cm]{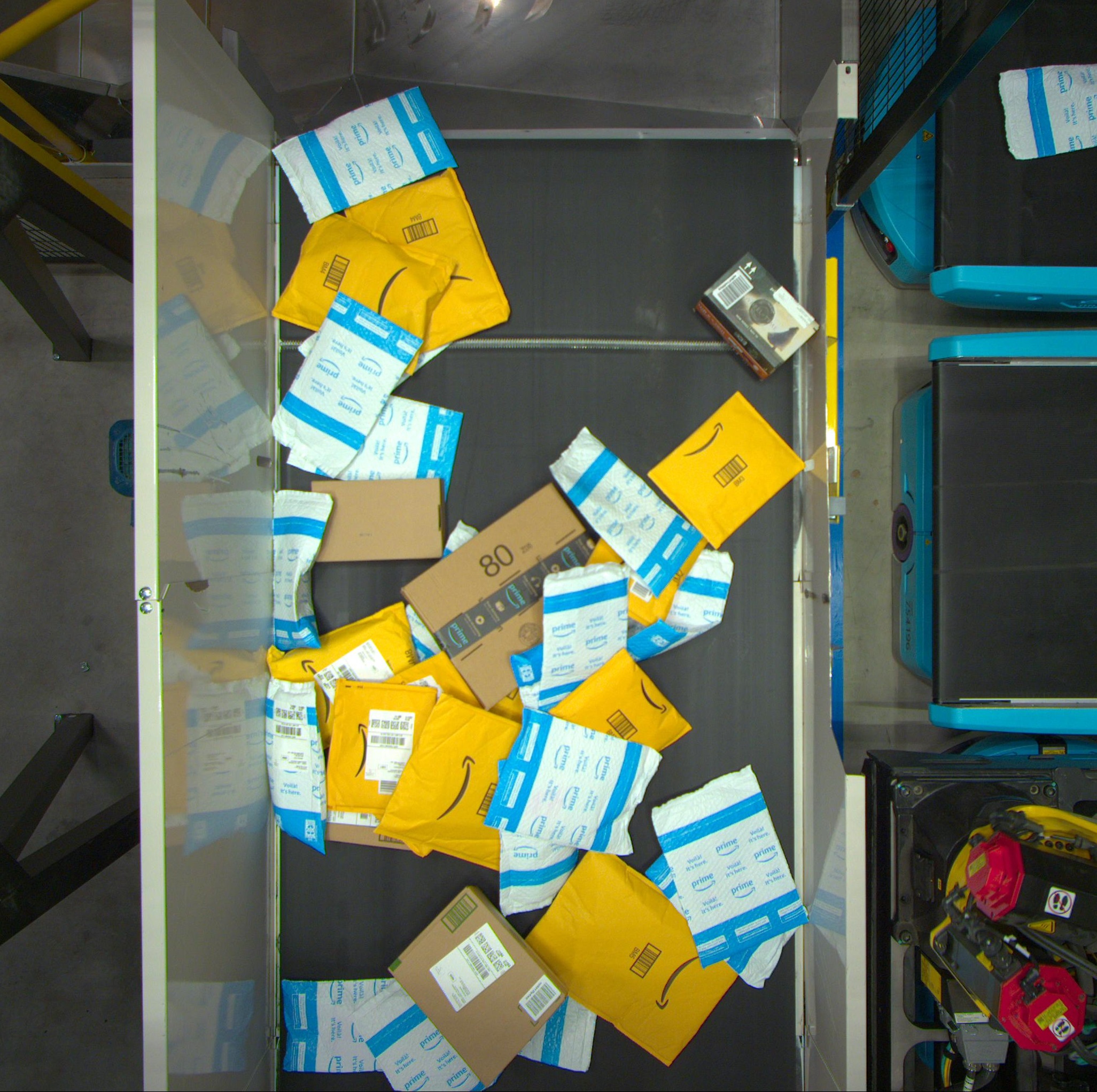}%
    \caption{A robot induction (Robin) workcell used for the statistics of this demonstration. The robotic arm is used for automated package singulation by Amazon.com, Inc. It picks packages from an unstructured pile on a conveyor belt and places them on mobile drive robots~\cite{Li:2023:rss:pickranking}.}
    \vspace{-.2in}
    \label{fig:robin-workcell}
\end{figure}

In a typical pick planning scenario, the inputs (e.g., RGB and depth) are provided to the pick generation system, and the system computes a set of one or more picks (ordered or unordered) that are provided to the motion planning system for execution, after some additional viability checks (e.g., collision and reachability checks). The details of the provided output can vary between different systems but typically include the position of the desired pick (i.e., \textit{pick point}), the rotation of the end-of-arm tool (EoAT) at the pick point, and the angle of the EoAT axis at the pick point (i.e., \textit{approach angle}). There may be additional parameters depending on the application. On the other hand, the approach angle in many methods is deterministic and not computed or generated by the pick planner. 

There are two common approaches for computing the desired picks from the inputs: \textit{whole-space approach} and \textit{sampled-space approach}. 

In the whole-space approach, the whole pick space is implicitly or explicitly considered to generate the picks for execution by the motion planning system: either the viable (or optimal) picks are directly generated from the inputs or the heat map of the whole pick space is computed (e.g., by a deep network)~\cite{gg-cnn, single-shot, mahler2016dex, dex-net}. The latter may generate picks based on specific optimality criteria or just generate viable picks. Note that the pick space can be discrete and finite in many practical cases (e.g., when only point-cloud elements are considered with a deterministic pick approach strategy). Ideally, whole-space approaches can compute the optimal set of picks over the entire pick space.

On the other hand, in the sampled-space approach, the system first generates (i.e., samples) a set of picks from the pick space (e.g., randomly or pre-determined), then either chooses the viable ones or ranks them based on specific scoring criteria before passing them to the motion planning system~\cite{Li:2023:rss:pickranking, geometric}. The main difference with the whole-space approach is that the sampled-space approach samples just a subset of the pick space, which may result in a set of bad or sub-optimal picks, but allows working with high-dimensional and large pick spaces to reduce the computational burden of the whole-space approach. 

Section~\ref{sec:our-methods} provides an overview of the approaches we developed over time for large-scale package manipulation at Amazon fulfillment centers; Section~\ref{sec:tests} illustrates our experiments highlighting the discussed methods; finally, Section~\ref{sec:conclusion} discusses the open problems and conclusions.

\section{Our Methods} \label{sec:our-methods}

\begin{figure}[!htb]
\centering
    \includegraphics[width=0.65\linewidth]{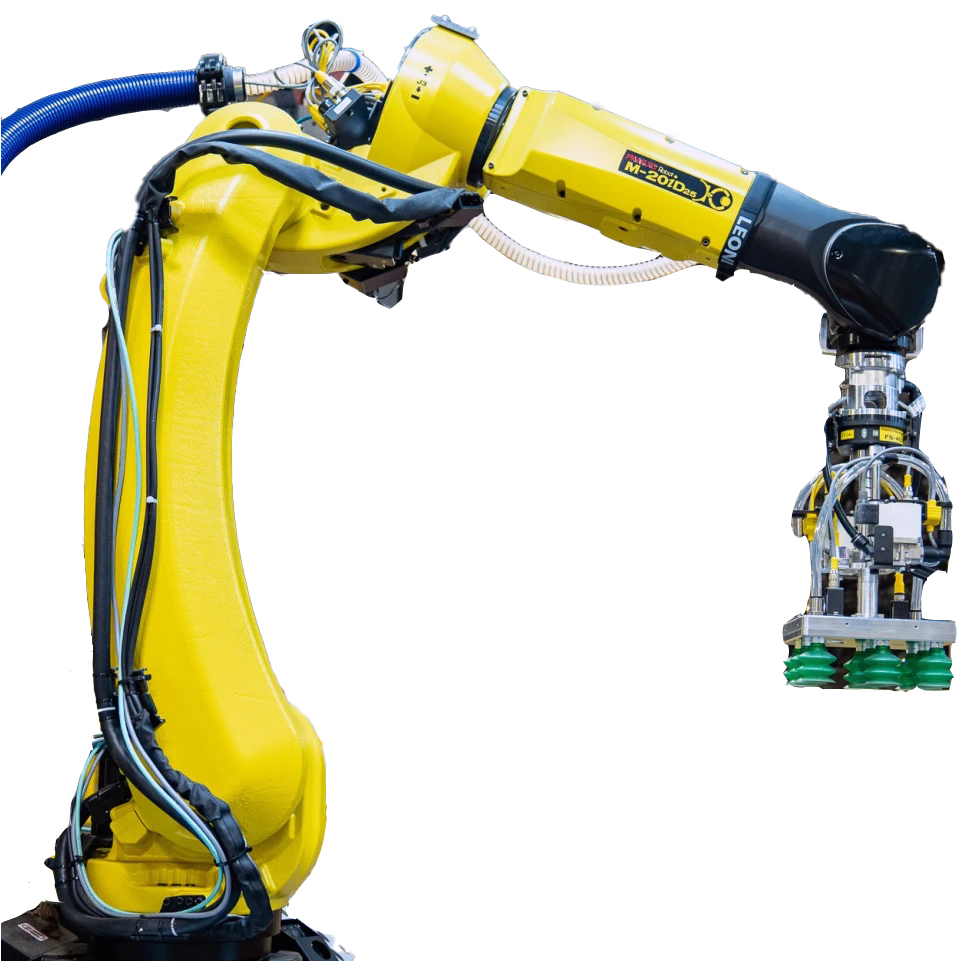}%
    \hfill%
    \includegraphics[width=0.3\linewidth]{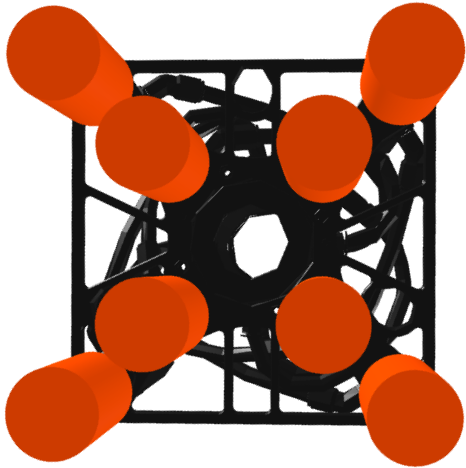}
    \caption{Robin robotic arm used in our experiments (left) and a simulation of the end-of-the-arm tool (EoAT) design (right)~\cite{Li:2023:rss:pickranking}.}
    \label{fig:robin}
    \vspace{-.1in}
\end{figure}

Figure~\ref{fig:robin-workcell} shows our Robin robots' typical scenario: a cluttered conveyor belt with various package types (deformable and rigid), different textures, and sizes. The robot needs to pick the packages one by one (i.e., singulate them) and place them on the mobile drives next to the conveyor belt. The robot's end-of-the-arm tool (EoAT) consists of a set of 8 suction cups that can be individually activated (see Figure~\ref{fig:robin}). Given the high dimension of the problem and strict required timing constraints for the whole operation, our solutions for large-scale manipulation have mainly focused on the sampled-space methods. 

Due to the workcell design, the pick planning system is provided with a top-view image of the workspace (i.e., the conveyor belt) similar to Figure~\ref{fig:robin-workcell}. All the packages in the image are segmented using a deep segmentation network based on Mask Scoring R-CNN~\cite{mask-scoring-rcnn} with a Swin-T backbone~\cite{swin-t}, which predicts the package material, instance segmentation masks, and the classification and segmentation scores. For rigid packages, we segment each surface visible in the image separately. For the rest of this document, we will refer to the segmented deformable packages and rigid package surfaces captured within the region of interest (ROI) of the workspace simply as \textit{segments}.

In our work, we started with the most straightforward heuristics for computing the order of the generated picks, then replaced them with more and more advanced algorithms over time. This section describes the methods used for generating the pick samples and then ranking them.

\subsection{Generating Picks}

Sampling a random pick with a uniform or a desired distribution over the pick space can cover the entire pick space given the large sample size and asymptotically will eventually pick the optimal pick. However, the sample size should be large to benefit from the advantages of random sampling. Given the timing constraints of the problem, processing to rank a large set of generated picks is not always viable. An alternative is to use a set of pre-defined samples more targeted toward choosing the feasible picks.

Given that most of the packages at Amazon have rectangular faces, in our development, we fit the largest inscribed ellipse into each segment in the image and look up the set of $N_p$ pre-computed picks for the resulting ellipse size. Each pick is for a different EoAT rotation and is the optimal center position with a set of activated cups with respect to the center of the fitted ellipse, where optimality is defined as the maximum number of suction cups fitting inside the ellipse with the center of their convex hull being closest to the center of the ellipse. Each generated pick is also associated with a \textit{quality score} computed based on the optimality criteria and read directly from the lookup table. Such a sampling strategy provides the desired number of reasonable picks very fast; however, it may miss the optimal picks, and it occasionally results in all picks failing to pass the viability tests, leaving the system without any remaining picks even when viable picks exist. 

When some of the $N_p$ picks do not pass a fast primitive collision check, we replace them with random picks (position and EoAT rotation) sampled from the pick space.

For each pick, the approach angle of the EoAT (i.e., the angle of EoAT's surface normal) is chosen based on the package material type: for packages with flat surfaces, the EoAT normal is aligned with the surface normal at the pick point, and for the rest, the approach angle is vertical (i.e., the EoAT will approach the pick point directly from above).

At the end of the pick generation step, for each segment seen in the top view RGB image, there will be $N_p$ (or less) picks generated that will be passed to the pick ranking algorithm.

\subsection{Ranking Segments}

Our system devises a two-step approach: first, it ranks the segments in the ROI, then it ranks the picks within each segment. The ordered list of picks is passed to the motion planning system for viability checks and execution. 

\begin{figure}[!htb]
\centering
    \begin{subfigure}[b]{.482\linewidth}
        \includegraphics[width=\linewidth]{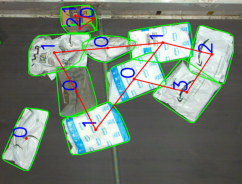}
        \caption{~}
        \label{fig:ranking:topo}
    \end{subfigure}%
    \hfill%
    \begin{subfigure}[b]{.50\linewidth}
        \centering
        \includegraphics[width=\linewidth]{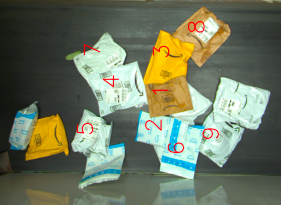}
        \caption{~}
        \label{fig:ranking:packages}
    \end{subfigure}%
    \caption{(a) Example of an adjacency graph. (b) Examples of ranking the packages based on their corresponding picks' highest success probability estimate. A smaller rank number represents a higher priority.}
    \label{fig:ranking}
\end{figure}

For \textit{ranking segments} for picking, over time, we developed several heuristic approaches, including:

\begin{itemize}[leftmargin=*]
    \item \textit{Package size:} Ranking the picks by the package size (or segmentation area). This heuristic assumes that picking packages with a larger visible area will have higher success and that moving out these larger packages first will help declutter the scene, making picking smaller packages easier. A major issue with this heuristic is when smaller packages lie on or overlap with larger ones. This can result in collisions with other packages, difficulty lifting a larger package from under smaller packages, or picking multiple packages simultaneously, resulting in failure to hold or singulate the target package.
    
    \item \textit{Z-order:} Ranking the picks by the package's target surface elevation. This heuristic is motivated by the assumption that a package whose surface is at the top of the pile is easily reachable, and the robot can avoid collisions or mistakenly pick other occluding packages when trying to pick it up. This heuristic is simple to implement but omits information about actual occlusions and, in practice, fails for unreachable picks and in instances where a portion of the package is at the top of the pile, but the rest is buried under other packages. In addition, since this method is only concerned with the elevation of the package's surface, it is not very informative for separate packages lying on the conveyor belt. 
    
    \item \textit{Topological order:} Ranking the picks based on the order of occlusion of the packages. Picks to package faces that appear unoccluded get the best score. Picks on package surfaces that are only occluded by unoccluded packages get the next highest score, and so on. This heuristic is moderately simple to implement and prioritizes the unoccluded packages to improve the success probability. However, it fails to recognize the unreachable package surfaces and the occluded packages with unoccluded surfaces, and similar to the Z-order method, it does not differentiate between the picks on the same package surface. Moreover, this method heavily relies on an accurate perception system to compute the segment overlaps. Figure~\ref{fig:ranking:topo} illustrates the topological graph computed for a pile of packages in the scene.
\end{itemize}

We also considered a combination of the above heuristics. For example, a topological order method is used as the primary ranking criteria, with the Z-order method as the tie-breaker when two packages have the same topological ranks. Some other heuristic approaches are also worth mentioning, such as the measure of how quadrilateral a package is or the confidence score given by the instance segmentation method. These approaches indirectly try to measure a package's occlusion or deformation level but have their own drawbacks and challenges. 

Our recent development replaced heuristics altogether with an online pick ranker that leverages the learned success predictor to prioritize the most promising picks for the robotic arm~\cite{Li:2023:rss:pickranking}. This learned ranking process is demonstrated to overcome the mentioned limitations and outperforms the performance of all the previous manually-engineered and heuristic alternatives. Figure~\ref{fig:ranking:packages} shows an example of segment ranking using this measure.

The developed learned pick quality measure ranks various pick alternatives in real-time and prioritizes the most promising ones for execution. The pick success predictor aims to estimate from prior experience the success probability of a desired pick by the deployed industrial robotic arms in cluttered scenes containing deformable and rigid objects with partially known properties. The success predictor employs a shallow machine learning model~\cite{catboost}, which allows evaluating which features are most important for success prediction. Li~et~al.~\cite{Li:2023:rss:pickranking} describes the details of the model and the training datasets collected to train the models.

\subsection{Ranking Picks Inside Segments}

For each segment, there are up to $N_p$ picks generated. Once the desired segment for picking is chosen, various strategies can be devised to order the candidate picks inside each segment. Some of these methods are based on simple, intuitive heuristics, such as preferring the picks closer to the center of the objects or the picks with more activated suction cups during EoAT's contact with the object. Therefore, historically, all our heuristic methods used the quality score provided by the pick generation system (which captures both metrics) to rank the picks inside each segment. 

In practice, such heuristics may work for a nominal induct but fail in complex scenarios and edge cases. Trying to manually handle all possible scenarios with more heuristics quickly becomes intractable. As an alternative, we use the score provided by the learned pick quality measure, which has the potential to capture the scenarios where the pre-defined quality score fails to consider.

\section{Experiments and Results} \label{sec:tests}

To compare the newly-developed methods with the deployed methods, at each step, we evaluated the performance of the two methods before and after the change. However, such a comparison is prone to biases and cannot reliably pick the better method when the differences are small (which is the case when the success rate of the methods reaches over 90\%). To confidently find the best strategy, a large-scale A/B test was deployed with six different experiment groups across the fleet, with a small percentage of total inducts allocated to each experiment group. Table~\ref{tbl:ab-large-results} summarizes the results of the A/B experiments. Please see Li~et~al.~\cite{Li:2023:rss:pickranking} for the detailed description of each experiment group.

\begin{table}[!htb]
   \caption{Results of A/B experiments for pick-ranking methods~\cite{Li:2023:rss:pickranking}.}
   \label{tbl:ab-large-results}
   \centering
   \begin{tabular}{lccc}
     \toprule
     Method & Total Picks & Failed Picks & Success Rate \\
     \midrule
     TopoZ-Center & 1,158,353 & 89,378 & 92.28\% \\
     Z-Center & 1,157,739 & 89,866 & 92.24\%\\
     TopoZ-Random & 1,158,479 & 109,193 & 90.57\%\\
    \cdashlinelr{1-4}
     TopoLPR-Center & 1,156,697 & 83,535 & 92.78\%\\
     LPR-Center & 1,160,005 & 72,789 & 93.73\%\\
     LPR-Random & 1,157,342 & 79,820 & 93.10\%\\
     \bottomrule
   \end{tabular}
\end{table}

The results show that using the learned pick success estimation to rank the segments (i.e., TopoLPR-Center and LPR-Center) improves the pick success for the robots compared with the manual heuristic ranking methods (i.e., TopoZ-Center and Z-Center). Interestingly, the best improvement comes from the more aggressive approach where we directly apply the learned pick success estimation for ranking (LPR-Center). Given that the heuristic ranking methods heavily depend on the heights of the packages, this suggests that promoting high packages can lead to more challenging picks, such as tall but unstable packages. On the other hand, the pick success model considers the package height only as an input feature and additional information such as the package position, surface normal, and adjacency graph features. Therefore, the pick success model can reason better and deprioritize unstable packages. It is also worth noting that if the picks are chosen randomly, the pick success improvement from the heuristic ranking method (TopoZ-Random) to the learned pick success estimation ranking method (LPR-Random) is even more significant (i.e., from 90.57\% to 93.10\%).

The pick success estimation model deployed in these A/B experiments predicts the pick success probability by taking the average of the predictions from five CatBoost models. The number of trees in the five CatBoost models is 2236, 1069, 799, 1464, and 1208 respectively. For all five models, the depth of the trees is 6, and we used a learning rate of 0.05.

\section{Conclusion} \label{sec:conclusion}

This extended abstract presented the evolution of pick planning methods developed at Amazon Robotics to manipulate and singulate packages in Amazon fulfillment centers. The current iteration uses the learned pick metric to rank the generated picks and is used for picking up to 6~million packages daily. It is the first large-scale deployment of learned methods for grasp and pick planning to the best of the authors' knowledge. The performed A/B evaluation of the different methods is the first experiment comparing heuristic and learned methods for pick planning at a large scale and provides empirical solid proof of the learned method's superiority over the heuristic and hand-engineered methods.

We believe that the recent developments in vision transformers combined with a large amount of induction data from our robotic fleet can improve our image-based network and may provide more valuable image embeddings, enhancing the prediction quality of the overall method.

\addtolength{\textheight}{-12cm}   



\section*{Acknowledgment}

The work would not have been possible without the support of the wider Amazon Robotics team. More specifically, the authors would like to thank previous and current members of the Robin and Janus teams, who established the underlying technology, provided support, and helped shape and deploy these ideas. The authors would like to give special thanks to David Oreper, Shuai Han, Qiujie Cui, Mansour Ahmed, and Andrew Marchese, whose help was critical in the realization of this work.


\bibliographystyle{IEEEtran}
\bibliography{references}

\begin{thebibliography}{10}
\providecommand{\url}[1]{#1}
\csname url@samestyle\endcsname
\providecommand{\newblock}{\relax}
\providecommand{\bibinfo}[2]{#2}
\providecommand{\BIBentrySTDinterwordspacing}{\spaceskip=0pt\relax}
\providecommand{\BIBentryALTinterwordstretchfactor}{4}
\providecommand{\BIBentryALTinterwordspacing}{\spaceskip=\fontdimen2\font plus
\BIBentryALTinterwordstretchfactor\fontdimen3\font minus
  \fontdimen4\font\relax}
\providecommand{\BIBforeignlanguage}[2]{{%
\expandafter\ifx\csname l@#1\endcsname\relax
\typeout{** WARNING: IEEEtran.bst: No hyphenation pattern has been}%
\typeout{** loaded for the language `#1'. Using the pattern for}%
\typeout{** the default language instead.}%
\else
\language=\csname l@#1\endcsname
\fi
#2}}
\providecommand{\BIBdecl}{\relax}
\BIBdecl

\bibitem{robin}
\BIBentryALTinterwordspacing
``{Amazon Robotics: Robin},'' 2022. [Online]. Available:
  \url{https://www.amazon.science/latest-news/robin-deals-with-a-world-where-things-are-changing-all-around-it}
\BIBentrySTDinterwordspacing

\bibitem{Li:2023:rss:pickranking}
\BIBentryALTinterwordspacing
S.~Li, A.~Keipour, K.~Jamieson, N.~Hudson, C.~Swan, and K.~Bekris,
  ``Demonstrating large-scale package manipulation via learned metrics of pick
  success,'' in \emph{Proceedings of Robotics: Science and Systems XIX (RSS
  2023)}, Daegu, Republic of Korea, 7 2023, pp. 1--11. [Online]. Available:
  \url{https://www.roboticsproceedings.org/rss19/p023.html}
\BIBentrySTDinterwordspacing

\bibitem{gg-cnn}
\BIBentryALTinterwordspacing
D.~Morrison, P.~Corke, and J.~Leitner, ``Learning robust, real-time, reactive
  robotic grasping,'' \emph{The International Journal of Robotics Research},
  vol.~39, no. 2-3, pp. 183--201, 2020. [Online]. Available:
  \url{https://journals.sagepub.com/doi/10.1177/0278364919859066}
\BIBentrySTDinterwordspacing

\bibitem{single-shot}
\BIBentryALTinterwordspacing
R.~Araki, T.~Hirakawa, T.~Yamashita, and H.~Fujiyoshi, ``{MT-DSSD: Multi-task}
  deconvolutional single shot detector for object detection, segmentation, and
  grasping detection,'' \emph{Advanced Robotics}, vol.~36, no.~8, pp. 373--387,
  2022. [Online]. Available:
  \url{https://www.tandfonline.com/doi/full/10.1080/01691864.2022.2043183}
\BIBentrySTDinterwordspacing

\bibitem{mahler2016dex}
\BIBentryALTinterwordspacing
J.~Mahler, F.~T. Pokorny, B.~Hou, M.~Roderick, M.~Laskey, M.~Aubry,
  K.~Kohlhoff, T.~Kr{\"o}ger, J.~Kuffner, and K.~Goldberg, ``{Dex-Net 1.0}: A
  cloud-based network of 3d objects for robust grasp planning using a
  multi-armed bandit model with correlated rewards,'' in \emph{IEEE
  International Conference on Robotics and Automation (ICRA)}, 2016, pp.
  1957--1964. [Online]. Available:
  \url{https://ieeexplore.ieee.org/document/7487342}
\BIBentrySTDinterwordspacing

\bibitem{dex-net}
\BIBentryALTinterwordspacing
J.~Mahler, M.~Matl, V.~Satish, M.~Danielczuk, B.~DeRose, S.~McKinley, and
  K.~Goldberg, ``Learning ambidextrous robot grasping policies,'' \emph{Science
  Robotics}, vol.~4, no.~26, p. eaau4984, 2019. [Online]. Available:
  \url{https://www.science.org/doi/abs/10.1126/scirobotics.aau4984}
\BIBentrySTDinterwordspacing

\bibitem{geometric}
\BIBentryALTinterwordspacing
A.~Morales, E.~Chinellato, A.~Fagg, and A.~del Pobil, ``Experimental prediction
  of the performance of grasp tasks from visual features,'' in
  \emph{Proceedings 2003 IEEE/RSJ International Conference on Intelligent
  Robots and Systems (IROS 2003) (Cat. No.03CH37453)}, vol.~4, 2003, pp.
  3423--3428 vol.3. [Online]. Available:
  \url{https://ieeexplore.ieee.org/document/1249685}
\BIBentrySTDinterwordspacing

\bibitem{mask-scoring-rcnn}
\BIBentryALTinterwordspacing
Z.~Huang, L.~Huang, Y.~Gong, C.~Huang, and X.~Wang, ``{Mask Scoring R-CNN},''
  in \emph{2019 IEEE/CVF Conference on Computer Vision and Pattern Recognition
  (CVPR)}, 2019, pp. 6402--6411. [Online]. Available:
  \url{https://ieeexplore.ieee.org/document/8953609}
\BIBentrySTDinterwordspacing

\bibitem{swin-t}
\BIBentryALTinterwordspacing
Z.~Liu, Y.~Lin, Y.~Cao, H.~Hu, Y.~Wei, Z.~Zhang, S.~Lin, and B.~Guo, ``Swin
  transformer: Hierarchical vision transformer using shifted windows,'' in
  \emph{2021 IEEE/CVF International Conference on Computer Vision (ICCV)},
  2021, pp. 9992--10\,002. [Online]. Available:
  \url{https://ieeexplore.ieee.org/document/9710580}
\BIBentrySTDinterwordspacing

\bibitem{catboost}
\BIBentryALTinterwordspacing
L.~Prokhorenkova, G.~Gusev, A.~Vorobev, A.~V. Dorogush, and A.~Gulin,
  ``{CatBoost}: unbiased boosting with categorical features,'' in
  \emph{Advances in Neural Information Processing Systems}, vol.~31.\hskip 1em
  plus 0.5em minus 0.4em\relax Curran Associates, Inc., 2018, pp. 1--11.
  [Online]. Available:
  \url{http://papers.neurips.cc/paper/7898-catboost-unbiased-boosting-with-categorical-features.pdf}
\BIBentrySTDinterwordspacing

\end{thebibliography}

\end{document}